%% file: ms.tex
\documentclass{article}
\usepackage{arxiv}

\usepackage[utf8]{inputenc}
\usepackage[ruled]{algorithm2e}

\usepackage{amsmath}
\usepackage[round]{natbib}
\usepackage{hyperref}
\usepackage{booktabs}
\usepackage{subcaption}
\usepackage[flushleft]{threeparttable}
\usepackage{graphicx}
\usepackage{afterpage}
\usepackage[super]{nth}
\newcommand{\ts}{\textsuperscript}
\usepackage[group-separator={,}]{siunitx}
\usepackage{pgfplots}
\pgfplotsset{width=4.5cm,compat=1.9}
\usetikzlibrary{arrows,positioning} 
\tikzset{
    >=stealth',
    node/.style={
           rectangle,
           rounded corners,
           font=\fontsize{8}{8.9}\selectfont,
           minimum height=1em,
           text centered},
    link/.style={
           thick,
           shorten <=2pt,
           shorten >=2pt,}
}

\begin{document}

\title{Neural-Network Guided Expression Transformation}         


\author{
  Romain Edelmann\\
  IC, EPFL\\
  Lausanne, Switzerland\\
  \texttt{romain.edelmann@epfl.ch} \\
   \And
 Viktor Kunčak \\
  IC, EPFL\\
  Lausanne, Switzerland \\
  \texttt{viktor.kuncak@epfl.ch} \\
}

\maketitle

\begin{abstract}
Optimizing compilers, as well as other translator systems, often work by rewriting expressions according to equivalence preserving rules.
Given an input expression and its optimized form, finding the sequence of rules that were applied is a non-trivial task. Most of the time, the tools provide no proof, of any kind, of the equivalence between the original expression and its optimized form.
In this work, we propose to reconstruct proofs of equivalence of simple mathematical expressions, after the fact, by finding paths of equivalence preserving transformations between expressions.
We propose to find those sequences of transformations using a search algorithm, guided by a neural network heuristic.
Using a Tree-LSTM recursive neural network, we learn a distributed representation of expressions
where the Manhattan distance between vectors approximately corresponds to the rewrite distance
between expressions.
We then show how the neural network can be efficiently used to search for transformation paths, leading to substantial gain in speed compared to an uninformed exhaustive search.
In one of our experiments, our neural-network guided search algorithm is able to solve more instances with a 2~seconds timeout per instance than breadth-first search does with a 5~minutes timeout per instance.
\end{abstract}


\section{Introduction}

Our work on proving equivalence of expressions is motivated by program transformations
and theorem proving objectives.
Many translator systems, such as optimizing compilers or simplifiers, are based on the idea of generating equivalent expressions from a given input expression. The goal is generally to find expressions that are more efficiently executable than the original one, or have more desirable latent properties.
For instance, in~\citep{1386651}, the authors present SPIRAL, an optimizing compiler for digital signal processing programs which is based on the idea of generating equivalent algorithms and searching for the most efficiently executable one.
Similarly, in~\citep{darulova2013synthesis}, the authors describe a method for compiling real-valued arithmetic expressions to fixed-point programs in a way that minimizes errors due to the loss of precision.
Their method is based on rewriting expressions based on rules which ensure the equivalence of the generated expressions on the real domain.
In~\citep{klonatos2013automatic} the authors describe a compilation scheme for memory hierarchy oblivious programs to memory hierarchy aware programs that also rely on expression transformations.
Often times, no explicit proof of equivalence between input and output is returned by the optimizing translators.
Indeed, generating certificates of equivalence is an extra burden for the translator program. Having such proofs would greatly increases the confidence in the correctness of the results of translators. Proof-carrying code~\citep{necula1998compiling} proposes generating proofs alongside translations. This approach has been
difficult to implement in practice.
The source code of the translator might not even be available in the first place, thereby greatly complicating any attempts at retrofitting proof-generation capabilities to the translator.
Finally, even though
randomized equivalence testing algorithms exist for certain
classes of expressions, the golden standard for
correctness is to find a sequence of sound rules that transform one expression
into another. More broadly, expression equivalence questions frequently arise
in interactive proof assistants such as Coq and Isabelle. 

In this work we aim to generate equivalence certificates in a way that is independent from the way the expressions are generated, and as such can be implemented as an external tool, independent of the original translator program. 
We propose to use paths of equivalence preserving transformations as certificates of  expressions equivalence.
Given such a path, the equivalence between two expressions can be easily checked by applying all transformations onto the first expression; if the resulting expression is syntactically equal to the second, then the proof is valid and the equivalence is witnessed.
In order for the proof to be easily comprehensible and checkable, we aim for the transformation paths to be short, but we impose no optimality constraint.

To find short paths between expressions, we propose to use a search algorithm. Using an uninformed exhaustive search, such as \emph{breadth-first search} (BFS), works well until the paths reach a certain size, after which the exponential size of the search space prohibits this approach.
Informed search algorithms, such as A*~\citep{hart1968formal}, are not directly applicable to the problem due to the lack of suitable heuristics for estimating the distance between expressions.

In this paper, we build a heuristic to approximate the distance between simple mathematical expressions in terms of number of equality preserving transformations (such as commutativity, associativity and distributivity). We use a recursive neural network, Tree-LSTM~\citep{tai2015improved}, to embed the mathematical expressions in a vector space where the Manhattan distance between vectors approximately corresponds to the distance between expressions. We use a supervised learning approach to train the network on a dataset of examples of pairs of expressions with their associated distance.

We develop two different pathfinding algorithms that make use of the trained neural network. The first algorithm, called \emph{Neural-Network Guided Search} (NNGS), explores the search tree in a best-first search manner --- nodes with a distributed representation closer to the distributed representation of the target will tend to be visited first.
The second algorithm, \emph{Batch Neural-Network Guided Search} (Batch-NNGS), is an adaptation of NNGS tailored to make use of the data-parallel processing capabilities of GPUs.


\section{Definitions}


\subsection{Expressions}

The expressions we consider are formed of variables, binary addition, binary multiplication, and a unique focus marker. The focus marker is a unary operator which indicates where transformations are to be applied. We denote the focus by $F({\cdot})$. We fix the maximum number of different variable names to 3. We use $a$, $b$ and $c$ as our 3 variables, and use the usual symbols for addition and multiplication.

\begin{figure}[h]
\centering{
\input{tikz/tree.tex}
\caption{Representation of the expression $a \cdot F(b + c)$ as a tree.}
\label{fig:tree}
}
\end{figure}
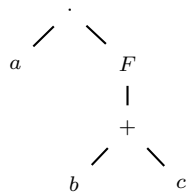

Expressions have a tree-like structure. Variables are leafs of the expression tree, while unary and binary operators are inner nodes of the tree.
Figure~\ref{fig:tree} shows such an expression tree.

\subsection{Length and Height}

We define the \emph{length} of an expression to be the number of operators and variables in the expression. The length of an expression corresponds to the number of nodes in the expression tree. For instance, the length of the expression $a \cdot F(b + c)$ is $6$.

We define the \emph{height} of an expression to be the height of its expression tree, i.e., the length of the longest shortest path from the root to a leaf. The height of the expression $a \cdot F(b + c)$ is $3$, as can be seen from figure~\ref{fig:tree}.

\subsection{Transformations}

The equality preserving transformations we consider are commutativity, associativity and distributivity.
In addition, we also consider navigational transformations, which move the focus marker around the expression.
Table~\ref{tab:transformations} describes the formal rules of the different transformations.

\begin{table}[h]
\centering{
\begin{threeparttable}
\begin{tabular}{@{}lrcl@{}} \toprule
Transformation & & Rule & \\ \midrule
Commutativity  & $\mathcal{C}[F(e_1 \circ e_2)]$ & $\mapsto$ & $\mathcal{C}[F(e_2 \circ e_1)]$  \\ \addlinespace
Associativity & $\mathcal{C}[F((e_1 \circ e_2) \circ e_3)]$ & $\mapsto$ & $\mathcal{C}[F(e_1 \circ (e_2 \circ e_3))]$  \\
                     & $\mathcal{C}[F(e_1 \circ (e_2 \circ e_3))]$ & $\mapsto$ & $\mathcal{C}[F((e_1 \circ e_2) \circ e_3)]$  \\ \addlinespace
Distributivity & $\mathcal{C}[F(e_1 \cdot (e_2 + e_3))]$ & $\mapsto$ & $\mathcal{C}[F((e_1 \cdot e_2) + (e_1 \cdot e_3))]$  \\ 
                    & $\mathcal{C}[F((e_1 \cdot e_2) + (e_1 \cdot e_3))]$ & $\mapsto$ & $\mathcal{C}[F(e_1 \cdot (e_2 + e_3))]$   \\ \addlinespace
Focus Up & $\mathcal{C}[F(e_1) \circ e_2]$ & $\mapsto$ & $\mathcal{C}[F(e_1 \circ e_2)]$ \\
                & $\mathcal{C}[e_1 \circ F(e_2)]$ & $\mapsto$ & $\mathcal{C}[F(e_1 \circ e_2)]$ \\ \addlinespace
Focus Left & $\mathcal{C}[F(e_1 \circ e_2)]$ & $\mapsto$ & $\mathcal{C}[F(e_1) \circ e_2]$ \\ \addlinespace
Focus Right & $\mathcal{C}[F(e_1 \circ e_2)]$ & $\mapsto$ & $\mathcal{C}[e_1 \circ F(e_2)]$ \\
\bottomrule
\end{tabular}
\begin{tablenotes}
      \small
      \item In each rule, $\circ$ indicates either addition or multiplication.
	$e_1, e_2$ and $e_3$ are arbitrary expressions without a focus.
	$\mathcal{C}$ denotes an arbitrary context without a focus. 
    \end{tablenotes}
\end{threeparttable}
\bigskip
\caption{Transformations.}
\label{tab:transformations}
}
\end{table}

\subsection{Transformation Paths}

We define \emph{transformation paths}, or simply \emph{paths}, between two expressions as sequences of transformations (including navigational transformations) which, when applied sequentially to the first expression, result in the second. We denote by $\mathcal{P}(e_1, e_2)$ the set of all paths between the expressions $e_1$ and $e_2$. We denote by $|p|$ the size of a path $p$.

\subsection{Rewrite Distance}

We define the \emph{rewrite distance} between two expressions to be the length of the shortest path between the two expressions, i.e.,  the minimum number of transformations necessary to transform the first expression into the second. Note that the set of paths between two expressions might be empty. In this case, we say that the rewrite distance is infinite.

\begin{equation}
d_{\text{rewrite}}(e_1, e_2) = \begin{cases}
\min\limits_{p \in \mathcal{P}(e_1, e_2)} |p| & \text{if $\mathcal{P}(e_1, e_2) \neq \emptyset$} \\
\infty & \text{otherwise}
\end{cases}
\end{equation}



\section{Neural Network Architecture}
\label{sec:architecture}

Our goal is to find a representation of expressions which is amenable to fast computation of the rewrite distance, or at least a good enough approximation of it.
This approximation of the rewrite distance will be used to guide the search, as explained in later sections.

\subsection{Neural Network}

\begin{figure}[t]
\centering{
\includegraphics[width=0.6\textwidth]{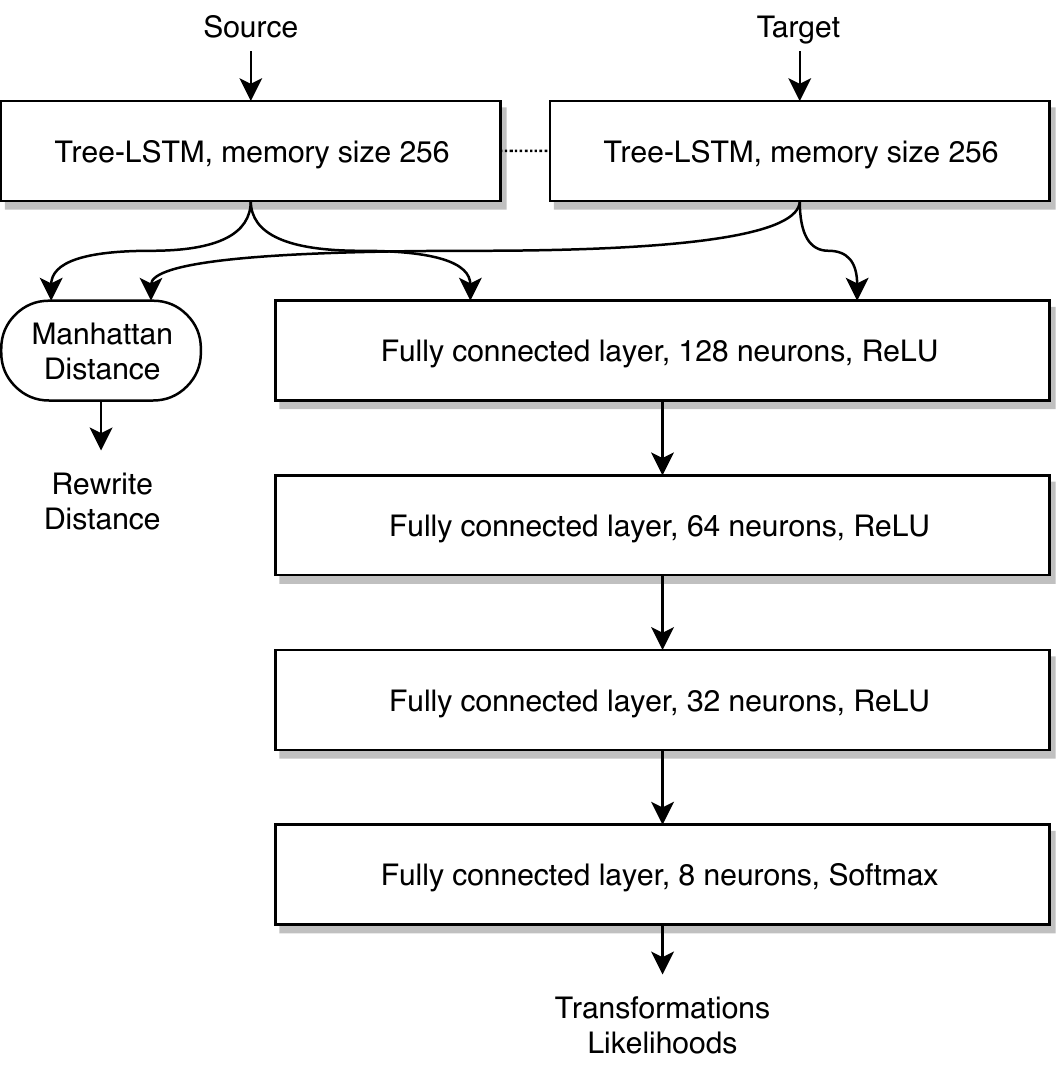}
\caption{The neural network architecture.}
\label{fig:network}
}
\end{figure}

Our key idea is to use a recursive neural network to embed expressions in a space where the Manhattan distance between vectors approximately corresponds to the rewrite distance between expressions. We use this recursive neural network as the surface layer of a larger network. The larger network takes as input two expressions and outputs a prediction of the rewrite distance, as well as a prediction of the first transformation to be applied. An overview of the architecture of the complete neural network is presented in figure~\ref{fig:network}.

\subsubsection{Inputs}
\label{sec:inputs}

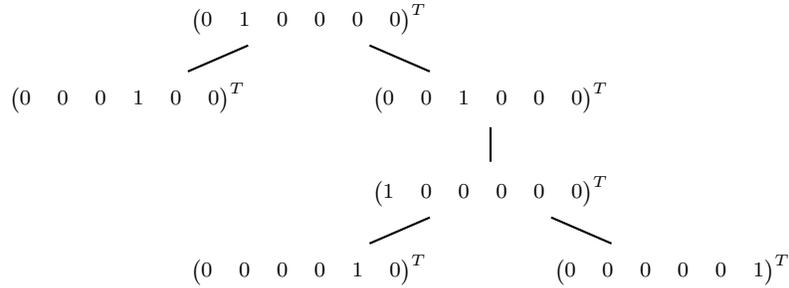
\begin{figure}[h]
\centering{
\input{tikz/tree-encoded.tex}
\caption{Representation of the expression $a \cdot F(b + c)$ as a tree, with one-hot encoding of the values.}
\label{fig:onehottree}
}
\end{figure}

The neural network receives as input two trees, one for each expression.
The values stored in the nodes of the trees are one-hot encodings of the corresponding variables or operators in the expression tree. Figure~\ref{fig:onehottree} shows the one-hot encoded expression tree corresponding to the expression $a \cdot F(b + c)$, whose tree is shown in figure~\ref{fig:tree}.

\subsubsection{Embedding}

At the surface of our neural network are two recursive subnetworks, whose objective is to represent the two tree-structured inputs as fixed-length vectors.
For this purpose, we use two Tree-LSTM~\citep{tai2015improved} subnetworks, each with a memory size of $256$.
The network is \emph{siamese}~\citep{bromley1994signature}, meaning that two Tree-LSTM subnetworks are identical, and share the same weights.
A discussion of Tree-LSTM appears later on in this paper.

\subsubsection{Rewrite Distance Prediction}

The Manhattan distance 
between the two $256$-dimension vectors obtained from the Tree-LSTM subnetworks is returned as an estimation of the rewrite distance between the two expressions.

\begin{equation}
\label{eq:manhattan}
\delta_{\text{Manhattan}}(v_1, v_2) = \lVert v_1 - v_2 \rVert_1 = \sum_{i} \lvert {v_1}_i - {v_2}_i \rvert
\end{equation}

The motivation behind the use of the Manhattan distance, or $L_1$ norm, comes the the fact that, as discussed by~\citet{aggarwal2001surprising}, this norm is appropriate in a high dimensional space, which is not the case for more "usual" norms such as $L_2$.

\subsubsection{First Transformation Prediction}

Underneath the two Tree-LSTM networks is also a series of fully connected layers composed of respectively 128, 64 and 32 neurons.
These fully connected layers use ReLU~\citep{nair2010rectified} as the activation function.
The last of these layers is fully connected to an 8 neurons layer using softmax as the activation function.
The output of this last layer serves as a prediction of the first transformation to be applied.

\begin{align}
\text{ReLU}(v)_i &= \left.
\begin{cases}
	v_i & \text{if $v_i >= 0$}\\
	0 & \text{otherwise}
\end{cases}
\right.
\label{eq:relu} \\
\text{softmax}(v)_i &= \frac{e^{v_i}}{\sum_j e^{v_j}} \label{eq:softmax}
\end{align}

\subsection{Tree-LSTM}

Tree-LSTM~\citep{tai2015improved} is a recursive neural network architecture inspired by the LSTM~\citep{hochreiter1997long} recurrent neural network architecture.
Contrarily to LSTM, which works on sequence-like inputs, Tree-LSTM works on tree-like inputs.
Since our expressions are tree-like structures, such recursive networks are a natural fit.

In recursive neural networks, the output of a given node depends on the input value associated with the node, as well as on the output values of the child nodes.
The network used to process each individual node is generally called the \emph{unit}.
Since the output of the network for a given node depends on the outputs of the children nodes, the input tree is processed in a bottom up fashion.

In our work, we use the \emph{$N$-ary Tree-LSTMs} variant of Tree-LSTM, as described in~\citep{tai2015improved}. This variant presupposes a maximal branching factor of $N$, which is $2$ in the case of our expressions. Its unit network is defined by the following equations:

\begin{align}
i_j &= \sigma \big( W^{(i)} x_j + \sum_{l=1}^N U_l^{(i)} h_{jl} + b^{(i)} \big) \label{eq:tree-lstm-i} \\
o_j &= \sigma \big( W^{(o)} x_j + \sum_{l=1}^N U_l^{(o)} h_{jl} + b^{(o)} \big) \label{eq:tree-lstm-o} \\
u_j &= \text{tanh} \big( W^{(u)} x_j + \sum_{l=1}^N U_l^{(u)} h_{jl} + b^{(u)} \big) \label{eq:tree-lstm-u} \\
f_{jk} &= \sigma \big( W^{(f)} x_j + \sum_{l=1}^N U_{kl}^{(f)} h_{jl} + b^{(f)} \big) \label{eq:tree-lstm-f} \\
c_j &= i_j \odot u_j + \sum_{l=1}^N \big( f_{jl} \odot c_{jl} \big) \label{eq:tree-lstm-c} \\
h_j &= o_j \odot \text{tanh}(c_j) \label{eq:tree-lstm-h}
\end{align}


In the above equations, $\odot$ denotes element-wise multiplication and $\sigma$ denotes the sigmoid function. 
The memory cell of the current node is denoted by $c_j$, while its hidden state is denoted by $h_j$.
The values $c_{jl}$ and $h_{jl}$ respectively denote the memory cell and the hidden state of the $l$\ts{th} child of the current node. 
The vector $x_j$ denotes the input value associated with the current node.

The parameter matrices $W^{(i)}$, $W^{(o)}$, $W^{(u)}$, and $W^{(f)}$ are of size $I \times M$, where $I$ is the dimension of the values and $M$ the memory dimension.
The parameter matrices $U_l^{(i)}$, $U_l^{(o)}$, $U_l^{(u)}$, and $U_{kl}^{(f)}$ for all $k, l \in \{ 1 \dots N \}$ are of size $M \times M$. The parameter bias vectors $b^{(i)}$, $b^{(o)}$, $b^{(u)}$, $b^{(f)}$ are each of size $M$.

The hidden state of the current node, as seen from equation~\ref{eq:tree-lstm-h}, is a partial view of the memory cell of the node.
The memory cell of the node directly depends on the values $i$ and $u$, as well as on the partially forgotten memory cells of child nodes (equation~\ref{eq:tree-lstm-c}).
The amount by which the memory cell of a child node is forgotten depends on the input value at the node, as well as on the hidden states of all children of the current node (equation~\ref{eq:tree-lstm-f}). We invite the readers to refer to~\citep{tai2015improved} for a more in-depth explanation.

In this work, the maximum branching factor $N$ is set to $2$, as the largest arity of operators in our expression language is $2$.
The input size $I$ is $6$ (one-hot encoding of addition, multiplication, focus marker and 3 variables). The memory dimension $M$ is set to $256$.

\subsection{Batch-Processing}

Batch-processing of recursive neural network is a non-trivial task~\citep{bowman2016fast}.
It is however highly desirable, as batch-processing exploits the data-parallel processing capabilities of specialized hardware, such as GPUs.
To enable batch processing of recursive neural networks, we used a variation of the technique found in~\citep{bowman2016fast}.
We released this work as an open source library for PyTorch~\citep{paszke2017automatic} called \texttt{treenet}.
The library is freely available online\footnote{The project is available at \url{https://github.com/epfl-lara/treenet}.}.

The key idea is to turn the recursive neural network into a recurrent neural network.
The tree-shaped input is turned into two sequences:
\begin{itemize}
\item a sequence of node values, and
\item a sequence of the corresponding node arities.
\end{itemize}
The values and arities of nodes are to appear in the two sequences in the order in which the nodes are visited by a \emph{post-order traversal} of the tree.
Note that no information is lost during this encoding process, as it is possible to reconstruct the original tree from its encoded representation.

\begin{figure}[t]
\centering{
\includegraphics[width=0.8\textwidth]{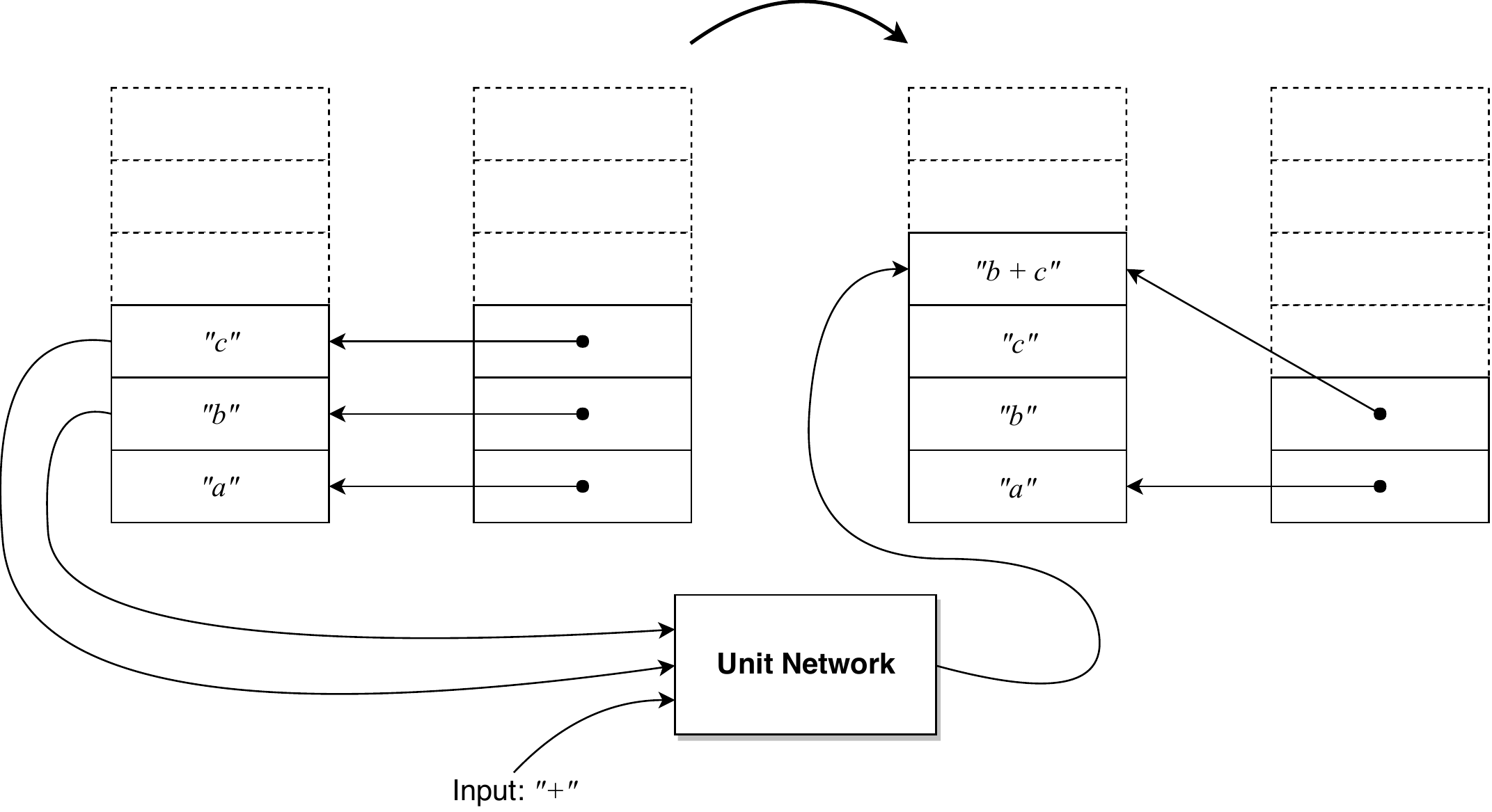}

\bigskip

\parbox{0.8\textwidth}{
As an example, we present one execution step that takes place during the processing of the input expression $a \cdot F(b + c)$.
Note that, in post-order traversal, the nodes are visited in the following order: $a, b, c, {+}, {F}, {\cdot}$.
At the point in time we consider, the distributed representations of the nodes $a$, $b$ and $c$ have already been computed and are sitting in the first stack, while the second stack maintains pointers to entries of the first. To process the next node, i.e., ${+}$, the distributed representations of $b$ and $c$ are retrieved from the first stack by following the top two pointers in the second stack. The two distributed representations, along with the one-hot encoded representation of the input ${+}$, are fed to the unit network (in the case of this work, a Tree-LSTM unit). Then, the resulting distributed representation is inserted into the first stack, while a pointer to it is inserted in the second stack, replacing the pointers to the arguments.
}
\caption{Example of one step of execution of the augmented recurrent neural network.}
\label{fig:stacks}
}
\end{figure}

The encoded input is fed to a recurrent neural network which, given the appropriate unit subnetwork, will emulate the execution of the corresponding recursive neural network.
The recurrent neural network maintains two stacks. The first stack is used to store the representation of the nodes (i.e., the outputs of the unit subnetwork), while the other stack maintains pointers to the first stack.
Given a tree encoded as post-order sequences of values and arities as input, the recurrent neural network proceeds, node by node, as follows:
\begin{enumerate}
\item
When a node is processed, the representation of all its children are retrieved. At this point, the pointers to the representations of its $n$ children, where $n$ is the recorded arity of the node, are found at the top of the second stack.
\item
The children representations, as well as the input associated with the current node, are passed down to the unit subnetwork, which in turn returns the distributed representation of the current node.
\item
The distributed representation of the current node is stored on top of the first stack.
\item
The $n$ pointers are removed from the top of the second stack, and replaced by a pointer to the newly inserted representation for the node.
\item
The next node in the sequence is processed.
\end{enumerate}
An example of one step of execution is presented in figure~\ref{fig:stacks}. 

The two stacks can trivially be turned data-parallel and implemented on GPUs, thereby enabling batch-processing of the recursive neural network. For batches containing sequences of varying number of nodes, the smaller sequences can be padded with special values. As this is simply a technical implementation detail, we do not discuss this point any further here.

\section{Training}

In this section, we discuss how the neural network was trained.
We first describe how we obtained the data used for training, validation and testing.
We then discuss the loss functions used.
Finally, we show the performance metrics for different epochs.

\subsection{Data}

Data used for training, validation and testing was generated from scratch.
We generated $56,763,200$ entries over the span of several weeks.
Each entry contains:
\begin{enumerate}
\item a source and a target expression,
\item the exact rewrite distance between the two expressions, and,
\item the first transformation on a shortest path from the source to the target. 
\end{enumerate}
The rewrite distances range from 1 to 10 inclusive.
The transformation considered are those presented in table~\ref{tab:transformations}.
The two \emph{Focus Up} transformations are grouped together, giving us a total of 8 different transformations.
We made sure every combination of rewrite distance and first transformation was represented equally often.

We randomly split the $\num{5763200}$ entries into $\num{5600000}$ training examples, $\num{160000}$ validation examples and $\num{3200}$ testing examples. The balance in rewrite distances and first transformation was preserved. Statistics on our data are presented in table~\ref{tab:data-stats}.

\begin{table}[h]
\centering{
\begin{tabular}{@{}lccc@{}} \toprule
	& Training & Validation & Testing \\ \midrule
Number of examples & $\num{5600000}$ & $\num{160000}$ & $\num{3200}$ \\ \addlinespace
Average length  & 21.08 & 26.60 & 33.42  \\ 
Minimum length & 8 & 10 & 18 \\
Maximum length & 186 & 126 & 104 \\ \addlinespace
Average height  & 5.25 & 5.61 & 6.04  \\ 
Minimum height & 3 & 3 & 4 \\
Maximum height & 12 & 11 & 11 \\
\bottomrule
\addlinespace
\end{tabular}
\caption{Data statistics.}
\label{tab:data-stats}
}
\end{table}


\subsection{Training}
\label{subsec:training}

The training loss was set to the sum of the \emph{mean squared error} of the rewrite distance prediction and the \emph{cross-entropy} of the transformation prediction, each discounted by the square root of the true rewrite distance.
The rationale behind this discount factor is that mistakes on easier problems (i.e., smaller rewrite distances) should be more heavily penalized than errors on more difficult problems.

\begin{align}
MSE(p, d) &= \sum_i \big( p_i - d_i \big)^2\\
CrossEntropy(t, c) &= - t_c + \log \Big( \sum_{i} \big( e^{t_i } \big) \Big) \\
Loss(p, t, d, c) &= \frac{MSE(p, d) + CrossEntropy(t, c)}{\sqrt{d}}
\end{align}

\bigskip

In the equations above, $p$ indicates the predicted rewrite distance, while $d$ indicates the true rewrite distance.
$t$ is the vector of transformations likelihoods, i.e., a unit vector of size corresponding to the number of transformations, with entries in $[0, 1]$. The value $c$ is the index of the actual transformation.

We trained on $\num{5600000}$ examples for 40 epochs and used a batch size of 128.
We used Adam~\citep{kingma2014adam} with default parameters as the optimization method.
Our implementation is in PyTorch~\citep{paszke2017automatic} and is freely available online\footnote{The project is available at \url{https://github.com/epfl-lara/nugget}.}.

The model metrics are shown in figure~\ref{fig:metrics}.
After 40 epochs of training, the mean absolute error of the rewrite distance on the testing set is $0.99$, while the accuracy of the prediction of first transformation is $78.47\%$.

\input{metrics.tex}

\clearpage

\section{Path Search}

As previously stated, our goal is to prove equality between expressions,
using concrete transformation paths as witnesses.
To prove the equality between two given expressions,
we perform a search from the first expression to the second.
We use the neural network described in the previous sections to guide the search.
We present two algorithms:
\begin{itemize}
\item \emph{Neural-Network Guided Search}, abbreviated NNGS, a search algorithm using the previously discussed neural network, and,
\item \emph{Batch Neural-Network Guided Search}, abbreviated Batch-NNGS, an adaptation of NNGS for leveraging the data-parallel computing capabilities of GPUs.
\end{itemize}

\subsection{Neural-Network Guided Search}

The first search algorithm we present is NNGS (see algorithm~\ref{alg:normal}).
NNGS maintains a priority queue of expressions to visit, ordered by increasing priority.
The priority of a node is the sum of the distance to the target expression, as estimated by the neural network, and a factor $\alpha$ times the distance from the source, or \emph{depth}, as seen during the search.

\begin{equation}
\label{eq:priority}
\text{priority}_{\mathit{depth}, \mathit{target}}(\mathit{expr}) = d_{\text{estimated}}(\mathit{expr}, \mathit{target}) + \alpha \cdot \mathit{depth}
\end{equation}

\bigskip

The parameter $\alpha$, usually set between $0$ and $1$, controls how much exploration is performed by the search algorithm. The larger the value of $\alpha$ is, the more the algorithm will tend to explore.

The neural network is also used to order transformations of expressions by likelihood.
Transformations with higher likelihood will be tried out sooner than expressions with lower likelihood.

The algorithm begins with only the $\mathit{source}$ expression in the priority queue.
At each loop iteration, the expression with smallest priority is retrieved, but not directly removed, from the queue.
The next most likely unvisited transformation of the expression is then retrieved.
If the transformation is the last untried transformation of the expression, the expression is removed from the (top of the) priority queue.
The transformation is then applied to the current expression.
If the resulting expression has already been visited, or if the transformation is invalid for the current expression, the loop proceeds directly to the next iteration.
Otherwise, the resulting expression is compared to the target expression.
In case they are equal, the algorithm returns the recorded path to the target and terminates.
Otherwise, the child expression is fed to the neural network to obtain an estimated distance to the target, as well as an ordering on its transformations.
The child expression is then inserted in the priority queue with the appropriate priority, as given by equation~\ref{eq:priority}.
The algorithm then proceeds with the next expression in the priority queue.

Note that, when using a parameter $\alpha$ strictly greater than $0$, the algorithm is bound to terminate if the target can be reached. Indeed, with $\alpha > 0$, the priority of a node tends to infinity as the depth of the node tends to infinity. This ensures that all nodes are eventually visited. This result however is more theoretical than practical, since in practice the search algorithms can time out, as well as run out of space.

\begin{algorithm}[t]
\SetAlgoNoLine
\SetKwFunction{FEmbedding}{ComputeEmbedding}
\SetKwFunction{FDistance}{EstimateDistance}
\SetKwFunction{FTransformations}{EstimateTransformationsLikelihoods}
\SetKwFunction{FInsert}{Insert}
\SetKwProg{Def}{Function}{ is}{end}
\KwIn{A source expression and a target expression}
\KwOut{A path between the source and the target expressions}
$\mathit{embeddingTarget}$ = \FEmbedding{$\mathit{target}$}\;
$\mathit{queue}$ = new priority queue\;

\Def{\FInsert{expr}}{
	$\mathit{embedding}$ = \FEmbedding{$\mathit{expr}$}\;
	$\mathit{distance}$ = \FDistance{$\mathit{embedding}$, $\mathit{embeddingTarget}$}\;
	$\mathit{likelihoods}$ = \FTransformations($\mathit{embedding}$, $\mathit{embeddingTarget}$)\;
	sort transformations of $\mathit{expr}$ according to $\mathit{likelihoods}$\;
	insert $\mathit{expr}$ into the $\mathit{queue}$ with priority $\mathit{distance}$ + $\alpha \cdot \text{depth of }\mathit{expr}$\;
}

\FInsert{$\mathit{source}$}\;

\While{$\mathit{target}$ is not reached and $\mathit{queue}$ is not empty}{
	$\mathit{current}$ = first expression in $\mathit{queue}$\;
	\eIf{all transformations of current have been visited}{
		remove $\mathit{current}$ from the $\mathit{queue}$\;
	}{
		$\mathit{transformation}$ = next unvisited transformation of $\mathit{current}$\;
		$\mathit{next}$ = apply $\mathit{transformation}$ to $\mathit{current}$\; 
		\If{$\mathit{next}$ is valid and has not yet been visited}{
			\If{$\mathit{next}$ is $\mathit{target}$}{
				\Return{path from $\mathit{source}$ to $\mathit{target}$}\; 
			}
			record depth of $\mathit{next}$ as depth of $\mathit{current} + 1$\;
			\FInsert{next}\;
		}	
	}
}
\caption{Neural-Network Guided Search}
\label{alg:normal}
\end{algorithm}

\clearpage

\subsection{Batch Neural-Network Guided Search}

The second algorithm we present is Batch-NNGS (see algorithm~\ref{alg:batched}).
While the NNGS search algorithm invokes the neural network one expression at a time, Batch-NNGS invokes the neural network in batches.
Invoking the neural network in batches leverages the data-parallel processing capabilities of GPUs, considerable reducing the running time of the algorithm in practice.
Batch-NNGS no longer makes use of the transformation likelihood estimations.

Batch-NNGS is an informed tree search algorithm that makes use of two queues:
\begin{itemize}
\item the \emph{main queue}, a \emph{first-in, first-out} queue,
\item the \emph{reserve queue}, a priority queue ordered by the priority previously described in equation~\ref{eq:priority}.
\end{itemize}
At each loop iteration, the first expression from the main queue is retrieved and all its valid and unseen neighbors are visited and inserted in the main queue. If the target expression is found amongst the child expressions, the recorded path to the target is returned and the algorithm terminates. This is similar to what is done with breadth-first search.

However, contrary to BFS, whenever the size of the main queue goes above a certain threshold (corresponding to the batch size), all its expressions are removed and fed to the Tree-LSTM network as a batch, thereby obtaining the distributed representations of all the expressions.
Then, the estimated rewrite distances to the target are obtained by computing the Manhattan distance between the distributed representations and the distributed representation of the target.
The expressions are then inserted in the \emph{reserve} queue, using the estimated distances, plus a factor $\alpha$ times their depth in the search tree, as their priority.

Afterwards, or whenever the main queue is empty, the top elements of the reserve queue are transferred to the main queue. The expressions are transferred from the reserve queue to the main queue until their priority is at least $1$ larger than the first, or a certain fixed number of elements have been transferred. We set this maximal number of transferred expressions to the batch size divided by the number of transformations. This ensures that all the transferred expressions can be processed before the main queue reaches its critical size and is emptied again. 

\begin{algorithm}[b]
\SetAlgoNoLine
\SetKwFunction{FEmbedding}{ComputeEmbedding}
\SetKwFunction{FEmbeddings}{ComputeEmbeddings}
\SetKwFunction{FDistances}{EstimateDistances}
\SetKwFunction{FTransformations}{EstimateTransformationsLikelihoods}
\SetKwFunction{FInsert}{Insert}
\SetKwProg{Def}{Function}{ is}{end}
\KwIn{A source expression and a target expression}
\KwOut{A path between the source and the target expressions}
$\mathit{embeddingTarget}$ = \FEmbedding{$\mathit{target}$}\;
$\mathit{mainQueue}$ = new FIFO queue\;
$\mathit{reserveQueue}$ = new priority queue\;
$K$ = $\mathit{batchSize} / \text{number of transformations}$\;

\While{$\mathit{target}$ is not reached and not ($mainQueue$ is empty and $reserveQueue$ is empty)}{
	\If{size of $\mathit{mainQueue} > \mathit{batchSize}$}{
		remove all elements from $\mathit{mainQueue}$ and name them $\mathit{exprs}$\;
		$\mathit{embeddings}$ = \FEmbeddings{$\mathit{exprs}$}\;
		$\mathit{distances}$ = \FDistances{$\mathit{embeddings, embeddingTarget}$}\;
		\For{each expr in exprs and corresponding distance in distances}{
			insert $\mathit{expr}$ into the $\mathit{reserveQueue}$ with priority $\mathit{distance}$ + $\alpha \cdot \text{depth of }\mathit{expr}$\;
		}
	}
	
	\If{mainQueue is empty}{
		\Repeat{priority larger than $1$ plus the first priority, and at most K times}{
			remove top element from $\mathit{reserveQueue}$ and insert it in $\mathit{mainQueue}$\;
		}
	}
	
	$\mathit{current}$ = first expression of $\mathit{mainQueue}$, removed\;
	\For{each transformation}{
		$\mathit{next}$ = apply $\mathit{transformation}$ to $\mathit{current}$\;
		\If{$\mathit{next}$ is valid has not yet been visited}{
			\If{$\mathit{next}$ is $\mathit{target}$}{
				\Return{path from $\mathit{source}$ to $\mathit{target}$}\; 
			}
			record depth of $\mathit{next}$ as depth of $\mathit{current} + 1$\;
			insert $\mathit{next}$ in $\mathit{mainQueue}$\;
		}
	}
}
\caption{Batch Neural-Network Guided Search}
\label{alg:batched}
\end{algorithm}

\clearpage

\section{Results}

We now evaluate the effectiveness of our method.
We compare both our algorithms to \emph{breadth-first search} (BFS), an uninformed exhaustive search.
All experiments were implemented in Python 3.5.4 with PyTorch 0.3.1 and CUDA 9.0.176. They were run on a server with 16 Intel® Xeon Gold 5122 3.60GHz CPUs, 131.7 GB of memory, and two NVIDIA® GeForce@ GTX 1080 Ti GPUs. The server is running Ubuntu 16.04.4 as its operating system.
While we had two GPUs available, our Batch-NNGS algorithm was restricted to only use one.
Both BFS and NNGS were run exclusively on CPUs.

\subsection{Effectiveness on Small Distances}

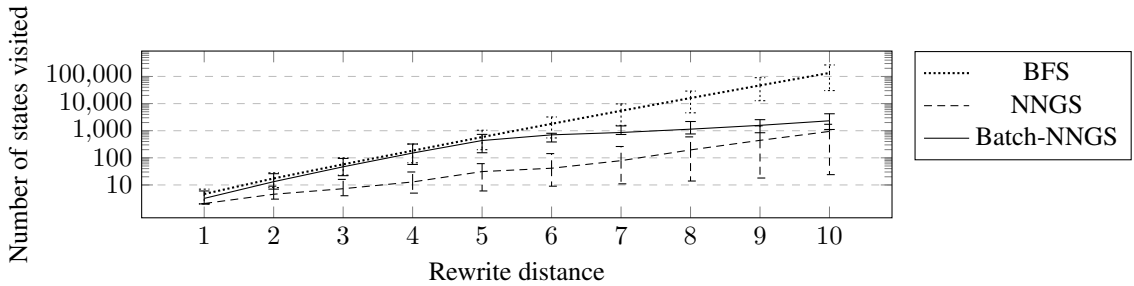
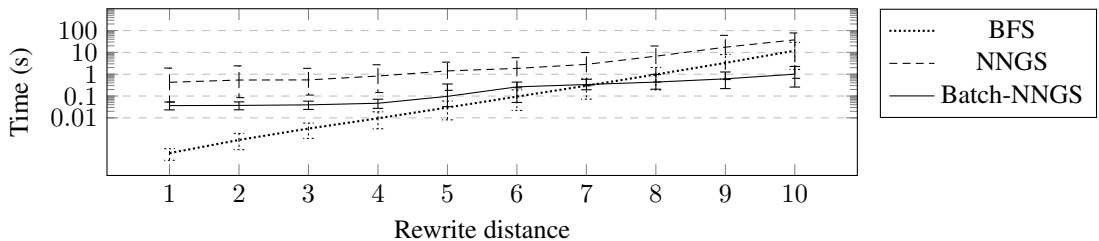
\begin{figure}[h]
\begin{subfigure}{\linewidth}
\centering{
\pgfplotsset{width=0.7\textwidth, height=3.8cm}
\begin{tikzpicture}
\begin{axis}[
	xlabel={Rewrite distance},
	ylabel={Number of states visited},
	legend pos=outer north east,
	ymode=log,
	ytick={10, 100, 1000, 10000, 100000},
	log ticks with fixed point,
	ymajorgrids=true,
    	grid style=dashed,
	]
\addplot+ [
		mark=none,
		black,
		densely dotted,
		thick,
                error bars/.cd,
                    y explicit,
                    y dir=both,
            ] table [
            	col sep=comma,
                x=distance,
                y=bfs_states_average,
                y error plus expr=\thisrow{bfs_states_95}-\thisrow{bfs_states_average},
                y error minus expr=\thisrow{bfs_states_average}-\thisrow{bfs_states_05},
            ] {data/search-small-states-256.csv};\addlegendentry{BFS}
            
\addplot+ [
		mark=none,
		black,
		densely dashed,
                error bars/.cd,
                    y explicit,
                    y dir=both,
            ] table [
            	col sep=comma,
                x=distance,
                y=nngs_states_average,
                y error plus expr=\thisrow{nngs_states_95}-\thisrow{nngs_states_average},
                y error minus expr=\thisrow{nngs_states_average}-\thisrow{nngs_states_05},
            ] {data/search-small-states-256.csv};\addlegendentry{NNGS}
            
\addplot+ [
		mark=none,
		black,
                error bars/.cd,
                    y explicit,
                    y dir=both,
            ] table [
            	col sep=comma,
                x=distance,
                y=batched_nngs_states_average,
                y error plus expr=\thisrow{batched_nngs_states_95}-\thisrow{batched_nngs_states_average},
                y error minus expr=\thisrow{batched_nngs_states_average}-\thisrow{batched_nngs_states_05},
            ] {data/search-small-states-256.csv};\addlegendentry{Batch-NNGS}
            
\end{axis}
\end{tikzpicture}
\caption{Average number of states visited by BFS, NNGS and Batch-NNGS on examples with a small rewrite distance. Error bars indicate the $0.95$ and $0.05$ percentiles.}
\label{fig:search-small-states}
}
\end{subfigure}
\par\bigskip 
\par\bigskip 
\begin{subfigure}{\linewidth}
\centering{
\pgfplotsset{width=0.7\textwidth, height=3.8cm}
\begin{tikzpicture}
\begin{axis}[
	xlabel={Rewrite distance},
	ylabel={Time (s)},
	ymax=1000,
	legend pos=outer north east,
	ymode=log,
	log ticks with fixed point,
	ytick={0,0.01,0.1,1,10,100},
	ymajorgrids=true,
    	grid style=dashed,
	]
\addplot+ [
		mark=none,
		densely dotted,
		thick,
		black,
                error bars/.cd,
                    y explicit,
                    y dir=both,
            ] table [
            	col sep=comma,
                x=distance,
                y=bfs_time_average,
                y error plus expr=\thisrow{bfs_time_95}-\thisrow{bfs_time_average},
                y error minus expr=\thisrow{bfs_time_average}-\thisrow{bfs_time_05},
            ] {data/search-small-time-256.csv};\addlegendentry{BFS}
            
\addplot+ [
		mark=none,
		densely dashed,
		black,
                error bars/.cd,
                    y explicit,
                    y dir=both,
            ] table [
            	col sep=comma,
                x=distance,
                y=nngs_time_average,
                y error plus expr=\thisrow{nngs_time_95}-\thisrow{nngs_time_average},
                y error minus expr=\thisrow{nngs_time_average}-\thisrow{nngs_time_05},
            ] {data/search-small-time-256.csv};\addlegendentry{NNGS}
            
\addplot+ [
		mark=none,
		black,
                error bars/.cd,
                    y explicit,
                    y dir=both,
            ] table [
            	col sep=comma,
                x=distance,
                y=batched_nngs_time_average,
                y error plus expr=\thisrow{batched_nngs_time_95}-\thisrow{batched_nngs_time_average},
                y error minus expr=\thisrow{batched_nngs_time_average}-\thisrow{batched_nngs_time_05},
            ] {data/search-small-time-256.csv};\addlegendentry{Batch-NNGS}
            
\end{axis}
\end{tikzpicture}
\caption{Average running time of BFS, NNGS and Batch-NNGS on examples with a small rewrite distance. Error bars indicate the $0.95$ and $0.05$ percentiles.}
\label{fig:search-small-time}
}
\end{subfigure}
\caption{Effectiveness of BFS, NNGS and Batch-NNGS on rewrite distances 1 to 10.}
\label{fig:search-small}
\end{figure}

As a first experiment, we compare the performance of NNGS, Batch-NNGS and breadth-first search on the $\num{3200}$ examples of our testing dataset. In this dataset, the rewrite distances range from 1 to 10. The batch size used for Batch-NNGS was set to $512$. The depth penalty parameter $\alpha$ was set to $0.5$ for both NNGS and Batch-NNGS. The results are reported in figure~\ref{fig:search-small}.

NNGS explored the least number of states, closely followed by Batch-NNGS. As expected, BFS explored the most number of states on average.
NNGS is however the slowest search on this dataset. This poor result can be explained by the fact that the algorithm invokes the neural network at each node, which is a relatively expensive operation.

Starting from rewrite distance $7$, Batch-NNGS is the fastest search algorithm.
Batch-NNGS is able to make use of the neural network to effectively prune the search space,
while making efficient use of the batch-processing capabilities of the GPU to reduce the cost of invoking the neural network.

\clearpage

\subsection{Effectiveness on Larger Distances}

\begin{figure}[h]
\centering{
\pgfplotsset{width=0.7\textwidth, height=5cm}
\begin{tikzpicture}
\begin{axis}[
	xlabel={Timeout per instance (s)},
	ymax=650,
	xmin = -10,
	xmax = 130,
	ylabel={Number of solvable instances},
	ytick={0, 100, 200, 300, 400, 500, 600},
	legend pos=outer north east,
	ymajorgrids=true,
    	grid style=dashed,
	]
\addplot+ [
		mark=none,
		black,
		thick,
		densely dotted,
            ] table [
            	col sep=comma,
                x=time,
                y=solved_bfs,
            ] {data/search-large-256.csv};\addlegendentry{BFS}
            
\addplot+ [
		mark=none,
		black,
		densely dashed,
            ] table [
            	col sep=comma,
                x=time,
                y=solved_nngs,
            ] {data/search-large-256.csv};\addlegendentry{NNGS}
            
\addplot+ [
		mark=none,
		black,
            ] table [
            	col sep=comma,
                x=time,
                y=solved_batched_nngs_512,
            ] {data/search-large-256.csv};\addlegendentry{Batch-NNGS}
            
\addplot+ [mark=none, loosely dashed
            ] coordinates {(-100, 600) (400, 600)};
            
\end{axis}
\end{tikzpicture}
\caption{Number of instances solvable by BFS, NNGS and Batch-NNGS under specified time. In this experiment, rewrite distances are larger than 10 and upper bounded by 200. The top dashed line indicates the total number of instances (600). }
\label{fig:search-large}
}
\end{figure}
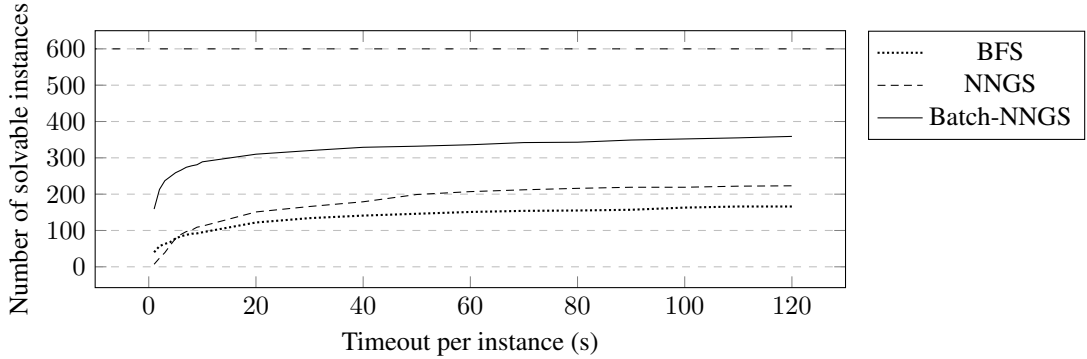

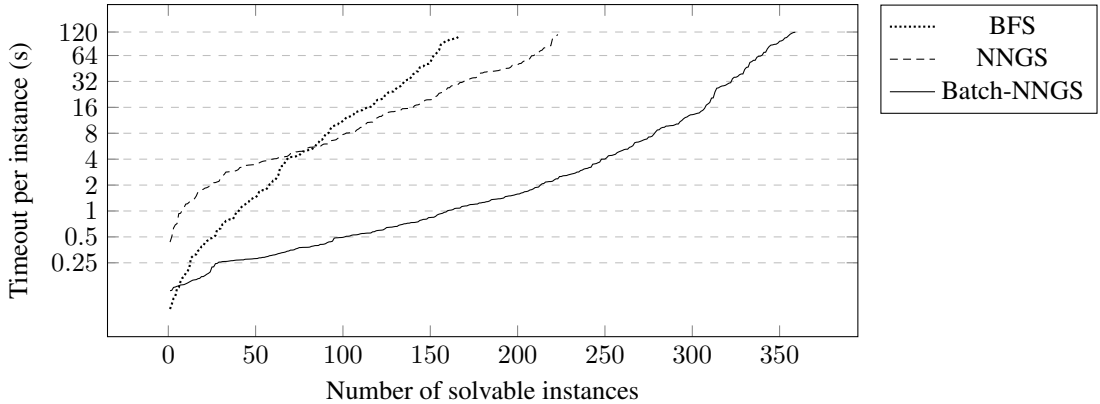
\begin{figure}[h]
\centering{
\pgfplotsset{width=0.7\textwidth, height=6cm}
\begin{tikzpicture}
\begin{axis}[
	xlabel={Number of solvable instances},
	ylabel={Timeout per instance (s)},
	legend pos=outer north east,
	ymajorgrids=true,
    	grid style=dashed,
	ymode=log,
	ytick={0.25, 0.5, 1, 2, 4, 8, 16, 32, 64, 120},
	log ticks with fixed point
	]
\addplot+ [
		mark=none,
		black,
		thick,
		densely dotted,
            ] table [
            	col sep=comma,
                x=number_solved,
                y=time_bfs,
            ] {data/search-large-transposed-256.csv};\addlegendentry{BFS}
            
\addplot+ [
		mark=none,
		black,
		densely dashed,
            ] table [
            	col sep=comma,
                x=number_solved,
                y=time_nngs,
            ] {data/search-large-transposed-256.csv};\addlegendentry{NNGS}
            
\addplot+ [
		mark=none,
		black,
            ] table [
            	col sep=comma,
                x=number_solved,
                y=time_batched_nngs_512,
            ] {data/search-large-transposed-256.csv};\addlegendentry{Batch-NNGS}

\end{axis}
\end{tikzpicture}
\caption{Time under which the specified number of instances can be solved.}
\label{fig:search-large-transposed}
}
\end{figure}

As a second experiment, we compare the performance of the NNGS, Batch-NNGS and BFS algorithms on random problems with larger rewrite distances.
To generate the random instances, we generated random source expressions of depth ranging from 4 to 5.
We obtained the targets by applying up to 200 random transformations to the sources.
We selected the first 600 entries where the rewrite distance was strictly larger than 10, witnessed by full exploration of the search space up to distance 10. 
We used the same parameters for NNGS and Batch-NNGS as in the previous experiment, i.e., a parameter $\alpha$ set to 0.5 for both NNGS and Batch-NNGS and a batch size of 512 for Batch-NNGS.
The results are presented in figures~\ref{fig:search-large}~and~\ref{fig:search-large-transposed}.

The Batch-NNGS algorithm is the most performant on this experiment. 
More instances were solved by Batch-NNGS within their first 2 seconds of execution than by BFS with a 5 minutes timeout per instance (213~vs~182).
Using a 2 minutes timeout per instance, Batch-NNGS solved 359 instances out of 600, while NNGS solved  223 and BFS only 166.

\subsection{Path Length}

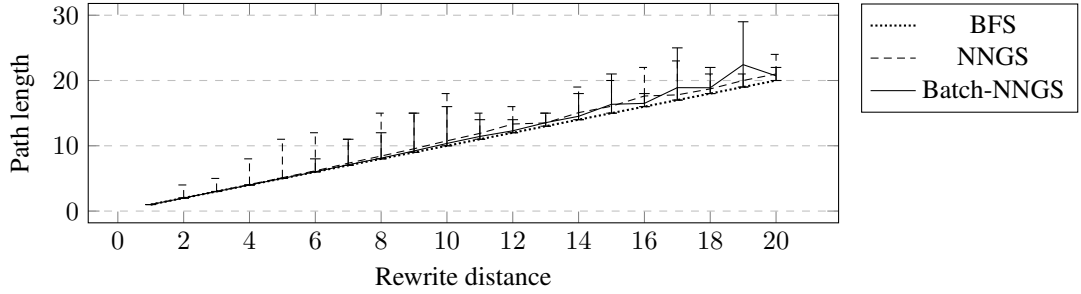
\begin{figure}[h]
\centering{
\pgfplotsset{width=0.7\textwidth, height=4.5cm}
\begin{tikzpicture}
\begin{axis}[
	xlabel={Rewrite distance},
	ylabel={Path length},
	legend pos=outer north east,
	ymajorgrids=true,
    	grid style=dashed,
	]
	
\addplot+ [
		mark=none,
		thick,
		black,
		densely dotted,
            ] table [
            	col sep=comma,
                x=bfs_distance,
                y=bfs_distance,
            ] {data/distances-256.csv};\addlegendentry{BFS}
            
\addplot+ [
		mark=none,
		black,
		densely dashed,
                error bars/.cd,
                    y explicit,
                    y dir=both,
            ] table [
            	col sep=comma,
                x=bfs_distance,
                y=nngs_distance_average,
                y error plus expr=\thisrow{nngs_distance_max}-\thisrow{nngs_distance_average},
                y error minus expr=\thisrow{nngs_distance_average}-\thisrow{nngs_distance_min},
            ] {data/distances-256.csv};\addlegendentry{NNGS}
            
\addplot+ [
		mark=none,
		black,
                error bars/.cd,
                    y explicit,
                    y dir=both,
            ] table [
            	col sep=comma,
                x=bfs_distance,
                y=batched_nngs_distance_average,
                y error plus expr=\thisrow{batched_nngs_distance_max}-\thisrow{batched_nngs_distance_average},
                y error minus expr=\thisrow{batched_nngs_distance_average}-\thisrow{batched_nngs_distance_min},
            ] {data/distances-256.csv};\addlegendentry{Batch-NNGS}
\end{axis}
\end{tikzpicture}
\caption{Average path length per rewrite distance. Error bars indicate the minimum and maximum value for the path lengths.}
\label{fig:distances}
}
\end{figure}

In the figure~\ref{fig:distances} we report the length of the paths found by NNGS, Batch-NNGS and BFS for rewrite distances 1 to 20.
The data for rewrite distances 1 to 10 come from the first experiment,
while the data for distances 11 to 20 come from the second experiment.

Since BFS performs an exhaustive search by increasing rewrite distance, it is guaranteed to always find paths of optimal length.
In the case of NNGS and Batch-NNGS, we have no such guarantees.
Indeed, in some encountered case, the paths found by NNGS and Batch-NNGS were longer than optimal.
However, the average length of paths found by NNGS and Batch-NNGS is close to the optimal on both experiments.

\subsection{Batch-NNGS Batch Size}

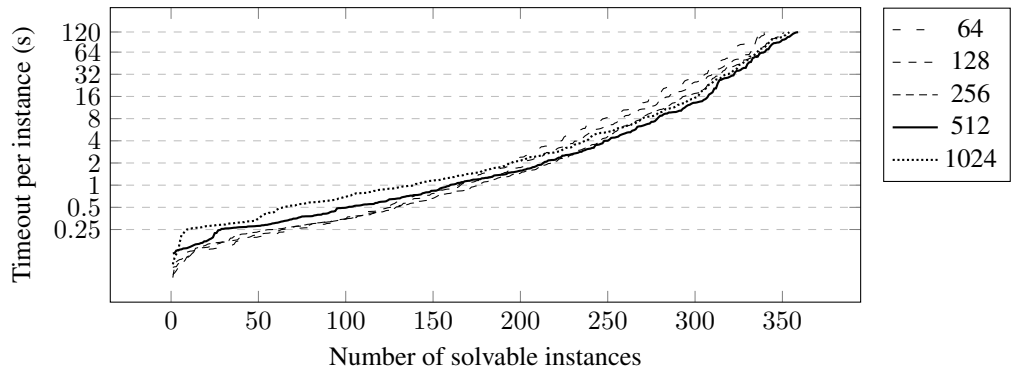
\begin{figure}[h]
\centering{
\pgfplotsset{width=0.7\textwidth, height=5.5cm}
\begin{tikzpicture}
\begin{axis}[
	xlabel={Number of solvable instances},
	ylabel={Timeout per instance (s)},
	legend pos=outer north east,
	ymajorgrids=true,
    	grid style=dashed,
	ymode=log,
	ytick={0.25, 0.5, 1, 2, 4, 8, 16, 32, 64, 120},
	log ticks with fixed point
	]
\addplot+ [
		mark=none,
		black,
		loosely dashed,
            ] table [
            	col sep=comma,
                x=number_solved,
                y=time_batched_nngs_64,
            ] {data/search-large-transposed-256.csv};\addlegendentry{64}
            
\addplot+ [
		mark=none,
		black,
		dashed,
            ] table [
            	col sep=comma,
                x=number_solved,
                y=time_batched_nngs,
            ] {data/search-large-transposed-256.csv};\addlegendentry{128}
            
\addplot+ [
		mark=none,
		black,
		densely dashed,
            ] table [
            	col sep=comma,
                x=number_solved,
                y=time_batched_nngs_256,
            ] {data/search-large-transposed-256.csv};\addlegendentry{256}
           
\addplot+ [
		mark=none,
		black,
		thick,
            ] table [
            	col sep=comma,
                x=number_solved,
                y=time_batched_nngs_512,
            ] {data/search-large-transposed-256.csv};\addlegendentry{512}
            
\addplot+ [
		mark=none,
		black,
		thick,
		densely dotted,
            ] table [
            	col sep=comma,
                x=number_solved,
                y=time_batched_nngs_1024,
            ] {data/search-large-transposed-256.csv};\addlegendentry{1024}
            
\end{axis}
\end{tikzpicture}
\caption{Time under which the specified number of instances can be solved depending on batch size.}
\label{fig:search-large-batches}
}
\end{figure}

In this experiment, we compare the performance of Batch-NNGS with batch sizes of 64, 128, 256, 512 and 1024, on the dataset of expressions with larger rewrite distances. The parameter $\alpha$ used was $0.5$. The results are reported in figure~\ref{fig:search-large-batches}. As seen from the graph, the performance of Batch-NNGS doesn't extremely vary with batch size. 
Using a small batch size (64, 128 or 256), Batch-NNGS is able to solve more instances using small timeout. Using a larger batch size (512 or 1024), Batch-NNGS is more efficient on longer timeouts.
In our experience, the parameter doesn't seem to have a significant influence on the average length of the returned paths.

\clearpage

\subsection{Batch-NNGS Parameter $\alpha$}

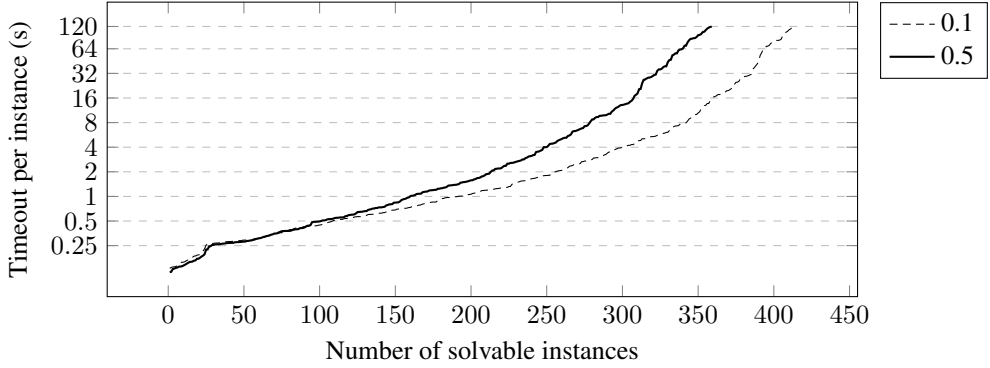
\begin{figure}[h]
\centering{
\pgfplotsset{width=0.7\textwidth, height=5.5cm}
\begin{tikzpicture}
\begin{axis}[
	xlabel={Number of solvable instances},
	ylabel={Timeout per instance (s)},
	legend pos=outer north east,
	ymajorgrids=true,
    	grid style=dashed,
	ymode=log,
	ytick={0.25, 0.5, 1, 2, 4, 8, 16, 32, 64, 120},
	log ticks with fixed point
	]
\addplot+ [
		mark=none,
		black,
		densely dashed,
            ] table [
            	col sep=comma,
                x=number_solved,
                y=time_batched_nngs_512_alpha_01,
            ] {data/search-large-transposed-256.csv};\addlegendentry{0.1}
            
%
\addplot+ [
		mark=none,
		black,
		thick,
            ] table [
            	col sep=comma,
                x=number_solved,
                y=time_batched_nngs_512,
            ] {data/search-large-transposed-256.csv};\addlegendentry{0.5}
            
            
\end{axis}
\end{tikzpicture}
\caption{Time under which the specified number of instances can be solved depending on the parameter $\alpha$.}
\label{fig:search-large-alpha}
}
\end{figure}

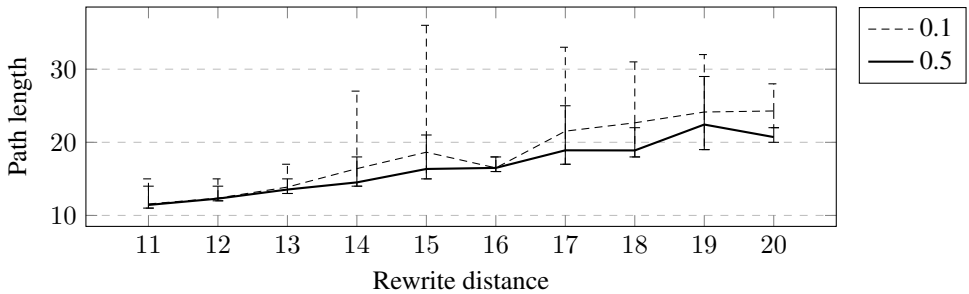
\begin{figure}[h]
\centering{
\pgfplotsset{width=0.7\textwidth, height=4.5cm}
\begin{tikzpicture}
\begin{axis}[
	xlabel={Rewrite distance},
	ylabel={Path length},
	legend pos=outer north east,
	ymajorgrids=true,
    	grid style=dashed,
	]
            
\addplot+ [
		mark=none,
		black,
		densely dashed,
                error bars/.cd,
                    y explicit,
                    y dir=both,
            ] table [
            	col sep=comma,
                x=distance,
                y=average_01,
                y error plus expr=\thisrow{max_01}-\thisrow{average_01},
                y error minus expr=\thisrow{average_01}-\thisrow{min_01},
            ] {data/distances-alphas.csv};\addlegendentry{0.1}
            
\addplot+ [
		mark=none,
		black,
		thick,
                error bars/.cd,
                    y explicit,
                    y dir=both,
            ] table [
            	col sep=comma,
                x=distance,
                y=average_05,
                y error plus expr=\thisrow{max_05}-\thisrow{average_05},
                y error minus expr=\thisrow{average_05}-\thisrow{min_05},
            ] {data/distances-alphas.csv};\addlegendentry{0.5}
\end{axis}
\end{tikzpicture}
\caption{Average path length per rewrite distance for different $\alpha$ parameters. Error bars indicate the minimum and maximum value for the path lengths.}
\label{fig:distances-alphas}
}
\end{figure}

In this experiment, we compare the performance of Batch-NNGS with parameter $\alpha$ of 0.1 and 0.5, on the dataset of expressions with larger rewrite distances used in the second experiment. The batch size used was 512. The results are reported in figures~\ref{fig:search-large-alpha}~and~\ref{fig:distances-alphas}. The parameter $\alpha$ is designed to regulate the amount of exploration performed by the algorithm. Using a small $\alpha$, Batch-NNGS performs less exploration and focuses more on the most promising nodes.
While this leads to better performance, the quality of the returned paths decreases. The average and maximum path length greatly increase when the parameter $\alpha$ is set to lower values.

\section{Future Improvements}

\subsection{Handling Arbitrary Number of Variables}

The neural network architecture discussed in the previous sections can only handle a fixed number of different variables. This is due to the encoding scheme presented in section~\ref{sec:inputs}. The fixed maximum number of variables encountered in expressions allows for a direct one-hot encoding of the operators and variables.
In this section, we discuss an alternative network architecture that we are experimenting with and that can handle expressions with an unbounded number of variables. The interface of the network remains unchanged --- it still takes as input a pair of expressions and returns an estimation of the rewrite distance and the predictions for the first transformation to apply.

The main idea behind the alternative architecture is to replace the one-hot encoding of operators and variables, which imposes a limit on the number of different variables, by learnt representations. The representation of variables depends on the input source and target expressions and must be computed for every different input expressions, whereas representation of operators is global.

The first phase in the new architecture consists in the computation of the representation of the variables appearing in the two input expressions.
In order to compute the representation of a given variable $x$, we first combine the two expression trees as a single tree rooted at specially marked node. We then apply a one-hot encoding to the values of the trees, with all variables different from $x$ sharing the same encoding. We then feed the encoded tree to a Tree-LSTM network. The resulting output vector is used as a the representation of the variable $x$. This process is reiterated for every different variables appearing in the source and target expressions.

Once the representations of the variables have been computed, encoded trees for the two input expressions are obtained by replacing every variables and operators in the expression trees by their representation. The rest of the network architecture --- as discussed in section~\ref{sec:architecture} --- is untouched.

The representation of variables and operators is learnt along with the representation of expressions in an end-to-end fashion. In our early experiments, after 30 epochs of training, the alternative architecture produced significantly better results than the previous architecture. On the testing set, the measured mean absolute error for the distance prediction is approximately $0.467$ (compared to $0.99$ previously) and the accuracy of the first transformation to be applied reaches $90\%$ (compared to $78.47\%$ previously).


\subsection{Use of Transformation Predictions in Batch-NNGS}

Another improvement that could be made is to incorporate the predictions of transformations in Batch-NNGS. Those transformation predictions are currently ignored by the algorithm. It is not yet clear how this information could be used.

\section{Discussion}

As seen from the results, our approach compares very favorably to uninformed exhaustive search.
Those results are encouraging and might suggest that the technique is applicable in practice.
We also expect to see large improvement in the running time of both our algorithms given dedicated hardware, such as FPGAs~\citep{8052266} or TPUs~\citep{jouppi2017datacenter}, for executing the neural network.

While we already obtained good results, we haven't been able to optimize for all the meta parameters given our finite resources.
Starting with the neural network model itself, which was only trained for a few days on a single machine.
We expect to see large improvements in the model quality given sufficient time and resources.
Also for this reason, we were only able to test a limited number of different neural network architectures and configurations in the few months that we worked on this project.
We were however able to see significantly better results using tree-structured networks compared to more traditional recurrent neural networks such as LSTM~\citep{hochreiter1997long}.

During the project, we also experimented with different sizes for the distributed representations vectors.
We initially set this size to 64, but obtained better results using 256.
As the number of weights in the LSTM unit is quadratic in the size of the distributed representations, we saw a large increase in the size of the model.
While this increased model size affects the performance of the algorithms,
the hit is compensated by the increased accuracy of the model.

Apart from the neural network model itself, our algorithms also have other meta parameters, which we could not completely optimize for. As seen from the last experiment, the parameter $\alpha$ has a great impact on the behavior of the algorithms. It is not clear to us which value should be set in practice.
Using a lower $\alpha$ results in shorter running times but in longer resulting path lengths.
This tradeoff between speed and quality of solutions has to be made on a case by case basis.
In some cases, it might be preferable to obtain good enough solutions quickly, while in some other cases, one might prefer to wait longer for results of better quality. 
The fact that this is parameterized, is, in our opinion, an advantage of the approach.

An other limiting factor is the availability of training data.
In our work, we generated our data from scratch --- a process that took several weeks.
Due to the large cost of getting accurate data for larger rewrite distance, we could only generate training examples up to rewrite distance 10.
Examples with larger rewrite distances take simply too long to generate.
We however were surprised to see that the algorithms were still able to work reasonably well on larger rewrite distances, even though examples with rewrite distances this large were not part of the training data.

\section{Related Work}

\subsection{The Knuth-Bendix Procedure}

The Knuth-Bendix completion~\citep{knuth1970simple} is a semi-algorithm for building a confluent set of directional rewriting rules from a set of equations.
Once the confluent set of rewriting rules is found, the task of checking, and proving, equality between two terms is trivially solved by applying the rules until both sides reach a normal form --- the two terms are semantically equal if and only if they reduce to the same normal form.
\citet{bachmair1989completion} present an extension of the Knuth-Bendix procedure with some termination guarantees. This extension is implemented in the Waldmeister~\citep{hillenbrand1997waldmeister} theorem prover.

The Knuth-Bendix procedure, and its extension, work fundamentally differently than our proposed approach.
The procedure first tries to find a confluent set of rewriting rules for a given set of equational axioms.
Then, if any such set is found, a path between any two equivalent expression can trivially be constructed by applying the constructed rewriting rules.
However, in the general case, there are no guarantees that the procedure will terminate and find a suitable set of rewrite rules.
Such a set of rewrite rules is not guaranteed to exist for any given domain.

\subsection{Pathfinding}

Pathfinding has a long history in the area of Artificial Intelligence.
Dijkstra's Algorithm~\citep{cormen2009introduction} is an exhaustive search procedure for weighted graphs.
The A* algorithm~\citep{hart1968formal} is an extension of Dijkstra's Algorithm that incorporates a heuristic function to guide the search. Assuming that the heuristic never overestimates distances, the A* algorithm is guaranteed to find the shortest path to the target.
Our proposed algorithms are also pathfinding algorithms.
Compared to the other discussed approaches, they however provide no guarantees of minimality of path found.

\subsection{Deep Reinforcement Learning and Monte-Carlo Tree Search}

In the area of Deep Reinforcement Learning and games, a number of advances have been made in the recent years. 
In~\citet{silver2016mastering} the authors present AlphaGo, a Go playing system which consistently beats top human and computer players.
In~\citep{silver2017zero}, the authors present an evolution of AlphaGo, named AlphaGo Zero, which doesn't use any human knowledge base and exceeds the performance of AlphaGo.
Finally, in~\citep{silver2017mastering}, the authors present an adaption of the technique to the games of Chess and Shogi.

The above systems are based on Monte-Carlo Tree Search (MCTS) algorithms~\citep{browne2012survey} and use neural network heuristics.
The goal of MCTS algorithms is to find moves that maximize a set reward function.
To do so, MCTS algorithms work iteratively in four phases:
\begin{description}
\item[Selection] A node of the search tree is selected for expansion. This selection depends on the average reward of the nodes and the number of times they were visited.
\item[Expansion] Child nodes, possibly several, are added to the selected node.
\item[Simulation] The reward of newly created nodes are estimated by random sampling of the search tree from that point.
\item[Backpropagation] The reward, and other statistics, of all parents of the newly expanded nodes are updated.
\end{description}
The neural network heuristics can be used to estimate the rewards of nodes and the frequency at which branches should be visited. In some cases, they also completely replace the simulation step.

On the surface, our approach to tree search and MCTS may seem very similar,
in the sense that both approaches use a neural network heuristic and tend to visit promising nodes first.
However, our approach does not take into account the frequency at which a node and its ancestors are currently visited to perform \emph{selection}, as we use other mechanisms to ensure exploration. We also do no perform random \emph{simulation}, and have no need for a \emph{backpropagation} phase.

\subsection{Distributed Representations for Natural Language Processing}

Distributed representations of words have been an instrumental part of many successes in the field of Natural Language Processing.
Word embeddings, such as Word2Vec~\citep{mikolov2013distributed} and GloVe~\citep{pennington2014glove} are able to represent words as fixed-length vectors capturing their semantics in great details. These methods, based on corpus statistics, are trained using unsupervised learning.
Distributed representations of words have been used to compute a semantic distance between text documents~\citep{kusner2015word}, in question answering~\citep{kumar2016ask} or for sentiment analysis~\citep{nakov2016semeval}, amongst many other examples. Distributed representation of sentences and text documents have also been successfully used, for instance in text classification~\citep{Le:2014:DRS:3044805.3045025}.
In~\citep{mueller2016siamese}, the authors learn a distributed representation of natural language sentences using a siamese LSTM~\citep{hochreiter1997long} architecture for text similarity. This approach is similar to the one used in this paper.

\subsection{Distributed Representations of Code}

Recently, distributed representations have been used in the context of code and programming languages.
\citet{allamanis2015suggesting} use a distributed representation of code to predict method and class names.
In~\citep{piech2015learning}, the authors learn a distributed representation of programs in order to give meaningful feedback to student programmers. 

\citet{2018arXiv180309473A} train a semantic capturing distributed representation of code. The authors then demonstrate that the distributed representation of code can be used to predict the name of methods based on their content with relatively high accuracy.

In the area of automated theorem proving, \citet{irving2016deepmath} train a distributed representation of logical formulas as part of a neural network used for premise selection.
In~\citep{2018arXiv180203685S}, the authors use a distributed representation of logical formulas as part of an experimental neural-network based SAT-solver.

%
%
%
%
%
%

\section{Conclusion}

We have shown that we can effectively search for transformations paths between simple mathematical expressions by using a neural network to guide the search.
As we only consider equality preserving transformations, the returned paths are concrete proofs that the two expressions are equivalent. The paths are generated completely independently of the tool that produced the expressions.

The neural network we have devised uses two identical Tree-LSTM recursive neural networks at its surface. The two Tree-LSTM subnetworks embed the tree-structured input expressions as vectors in a high dimensional space. The network is trained so that the Manhattan distance between distributed representations corresponds approximately to the rewrite distance between the corresponding expressions. In addition to returning the estimated distance between two expressions, the neural network also suggests the first transformation to be applied.

We have devised two algorithms which make use of the neural network.
The first, NNGS, makes use of the distance prediction outputted by the network, as well as of the outputted transformation predictions, to effectively prune the search space.
However, due to the costly invocation of the neural network performed at every single visited node in the search tree, the runtime performance of NNGS was unsatisfactory.

To have better runtime performance, we devised our second algorithm, Batch-NNGS, which queries the neural network in batches. While the number of states visited by Batch-NNGS tends to be larger than the number visited by NNGS, the modified algorithm is considerably faster.
This gap in performance compared to NNGS is explained by the fact that Batch-NNGS queries the neural network in batches --- an operation which can be efficiently performed on GPUs.

We have explored the different parameters of our algorithms and showed how they affect their behavior.
We have shown that the parameters of our algorithms can be tweaked for either performance, or quality of the returned paths.
While we have already obtained encouraging results,
we expect to see even better results given more accurate models or dedicated hardware.



\bibliography{bibliography}

%

\end{document}

%% file: tikz/tree.tex
\begin{tikzpicture}[node distance=0.4cm, auto,]
\node[node] (TIMES) {$\cdot$};
\node[below=of TIMES] (DUMMY_TIMES) {};
\node[node, left=of DUMMY_TIMES] (a) {$a$};
\node[node, right=of DUMMY_TIMES] (F) {$F$};
\node[node, below=of F] (PLUS) {$+$};
\node[below=of PLUS] (DUMMY_PLUS) {};
\node[node, left=of DUMMY_PLUS] (b) {$b$};
\node[node, right=of DUMMY_PLUS] (c) {$c$};

\draw (TIMES) edge [link] (a)
          (TIMES) edge [link] (F)
          (F) edge [link] (PLUS)
          (PLUS) edge [link] (b)
          (PLUS) edge [link] (c); 

\end{tikzpicture}

%% file: tikz/tree-encoded.tex
\begin{tikzpicture}[node distance=0.6cm, auto,]
\node[node] (TIMES) {$\begin{pmatrix} 0 & 1 & 0 & 0 & 0 & 0 \end{pmatrix}^T$};
\node[below=of TIMES] (DUMMY_TIMES) {};
\node[node, left=of DUMMY_TIMES] (a) {$\begin{pmatrix} 0 & 0 & 0 & 1 & 0 & 0 \end{pmatrix}^T$};
\node[node, right=of DUMMY_TIMES] (F) {$\begin{pmatrix} 0 & 0 & 1 & 0 & 0 & 0 \end{pmatrix}^T$};
\node[node, below=of F] (PLUS) {$\begin{pmatrix} 1 & 0 & 0 & 0 & 0 & 0 \end{pmatrix}^T$};
\node[below=of PLUS] (DUMMY_PLUS) {};
\node[node, left=of DUMMY_PLUS] (b) {$\begin{pmatrix} 0 & 0 & 0 & 0 & 1 & 0 \end{pmatrix}^T$};
\node[node, right=of DUMMY_PLUS] (c) {$\begin{pmatrix} 0 & 0 & 0 & 0 & 0 & 1 \end{pmatrix}^T$};

\draw (TIMES) edge [link] (a)
          (TIMES) edge [link] (F)
          (F) edge [link] (PLUS)
          (PLUS) edge [link] (b)
          (PLUS) edge [link] (c); 

\end{tikzpicture}

%% file: metrics.tex
\begin{figure}
\begin{subfigure}{0.5\linewidth}
\centering
\begin{tikzpicture}
\begin{axis}[
    xlabel={Epoch},
    ylabel={MAE},
    xmin=-1, xmax=42,
    ymin=0.8, ymax=1.5,
    xtick={1, 10, 20, 30, 40},
    ytick={0,0.2,0.4,0.6,0.8,1,1.2,1.4},
    legend pos=outer north east,
    ymajorgrids=true,
    grid style=dashed,
    mark repeat={3},
]

\addplot+[
    mark=none,
    densely dashed,
    black,
    ]
    table[x expr=\thisrowno{0} + 1,y=mae,col sep=comma]{data/training_256.csv};
    \addlegendentry{Training}
    
\addplot+[
    mark=none,
    densely dotted,
    thick,
    black,
    ]
    table[x expr=\thisrowno{0} + 1,y=mae,col sep=comma]{data/validation_256.csv};
    \addlegendentry{Validation}

\addplot+[
    mark=none,
    black,
    ]
    table[x expr=\thisrowno{0} + 1,y=mae,col sep=comma]{data/testing_256.csv};
    \addlegendentry{Testing}

\end{axis}
\end{tikzpicture}
\caption{Mean absolute error of edit distance}
\end{subfigure}%
\begin{subfigure}{0.5\linewidth}
\centering
\begin{tikzpicture}
\begin{axis}[
    xlabel={Epoch},
    ylabel={Accuracy (\%)},
    xmin=-1, xmax=42,
    ymin=0, ymax=100,
    xtick={1, 10, 20, 30, 40},
    ytick={0,20,40,60,80,100},
    legend pos=outer north east,
    ymajorgrids=true,
    grid style=dashed,
]
 
\addplot+[
    mark=none,
    densely dashed,
    black,
    ]
    table[x expr=\thisrowno{0} + 1,y expr=\thisrowno{2} * 100,col sep=comma]{data/training_256.csv};
    \addlegendentry{Training}

\addplot+[
    mark=none,
    densely dotted,
    thick,
    black,
    ]
    table[x expr=\thisrowno{0} + 1,y expr=\thisrowno{2} * 100,col sep=comma]{data/validation_256.csv};
    \addlegendentry{Validation}

\addplot+[
    mark=none,
    black,
    ]
    table[x expr=\thisrowno{0} + 1,y expr=\thisrowno{2} * 100,col sep=comma]{data/testing_256.csv};
    \addlegendentry{Testing}

\end{axis}
\end{tikzpicture}
\caption{Accuracy of first transformation prediction}
\end{subfigure}
\par\bigskip 
\par\bigskip 
\begin{subfigure}{0.5\linewidth}
\centering
\begin{tikzpicture}
\begin{axis}[
    xlabel={Epoch},
    ylabel={Discounted MSE},
    xmin=-1, xmax=42,
    ymin=0, ymax=1,
    xtick={1, 10, 20, 30, 40},
    ytick={0,0.2,0.4,0.6,0.8,1,1.2,1.4},
    legend pos=outer north east,
    ymajorgrids=true,
    grid style=dashed,
    mark repeat={3},
]
 
\addplot+[
    mark=none,
    densely dashed,
    black,
    ]
    table[x expr=\thisrowno{0} + 1,y=dmse,col sep=comma]{data/training_256.csv};
    \addlegendentry{Training}   
    
\addplot+[
    mark=none,
    densely dotted,
    thick,
    black,
    ]
    table[x expr=\thisrowno{0} + 1,y=dmse,col sep=comma]{data/validation_256.csv};
    \addlegendentry{Validation}

\addplot+[
    mark=none,
    black,
    ]
    table[x expr=\thisrowno{0} + 1,y=dmse,col sep=comma]{data/testing_256.csv};
    \addlegendentry{Testing}

\end{axis}
\end{tikzpicture}
\caption{Discounted MSE of edit distance}
\end{subfigure}%
\begin{subfigure}{0.5\linewidth}
\centering
\begin{tikzpicture}
\begin{axis}[
    xlabel={Epoch},
    ylabel={Discounted CE},
    xmin=-1, xmax=42,
    ymin=0, ymax=1,
    xtick={1, 10, 20, 30, 40},
    ytick={0,0.2,0.4,0.6,0.8,1,1.2,1.4},
    legend pos=outer north east,
    ymajorgrids=true,
    grid style=dashed,
    mark repeat={3},
]
 
\addplot+[
    mark=none,
    densely dashed,
    black,
    ]
    table[x expr=\thisrowno{0} + 1,y=dce,col sep=comma]{data/training_256.csv};
    \addlegendentry{Training}

\addplot+[
    mark=none,
    densely dotted,
    thick,
    black,
    ]
    table[x expr=\thisrowno{0} + 1,y=dce,col sep=comma]{data/validation_256.csv};
    \addlegendentry{Validation}

\addplot+[
    mark=none,
    black,
    ]
    table[x expr=\thisrowno{0} + 1,y=dce,col sep=comma]{data/testing_256.csv};
    \addlegendentry{Testing}

\end{axis}
\end{tikzpicture}
\caption{Discounted Cross Entropy}
\end{subfigure}
\par\bigskip 
\par\bigskip 
\begin{subfigure}{0.5\linewidth}
\centering
\begin{tikzpicture}
\begin{axis}[
    xlabel={Edit distance},
    ylabel={MAE},
    ymin= 0, ymax=2,
    ytick={0, 0.5, 1, 1.5, 2},
    xtick={1, 4, 7, 10},
    xmin=0, xmax=11,
    legend pos=outer north east,
    ymajorgrids=true,
    grid style=dashed,
]
 
\addplot+[
    mark=none,
    densely dashed,
    black,
    ]
    table[x=distance,y=mae,col sep=comma]{data/metrics_per_distance_training.csv};
    \addlegendentry{Training} 
    
\addplot+[
    mark=none,
    densely dotted,
    thick,
    black,
    ]
    table[x=distance,y=mae,col sep=comma]{data/metrics_per_distance_validation.csv};
    \addlegendentry{Validation} 
 
\addplot+[
    mark=none,
    black,
    ]
    table[x=distance,y=mae,col sep=comma]{data/metrics_per_distance_testing.csv};
    \addlegendentry{Testing}   

\end{axis}
\end{tikzpicture}
\caption{MAE of distance per edit distance (last epoch)}
\end{subfigure}%
\begin{subfigure}{0.5\linewidth}
\centering
\begin{tikzpicture}
\begin{axis}[
    xlabel={Edit distance},
    ylabel={Accuracy (\%)},
    xmin=0, xmax=11,
    ymin=0, ymax=110,
    ytick={0,20,40,60,80,100},
    xtick={1, 4, 7, 10},
    legend pos=outer north east,
    ymajorgrids=true,
    grid style=dashed,
]
 
\addplot+[
    mark=none,
    densely dashed,
    black,
    ]
    table[x=distance,y expr=\thisrowno{2} * 100,col sep=comma]{data/metrics_per_distance_training.csv};
    \addlegendentry{Training} 
    
\addplot+[
    mark=none,
    densely dotted,
    thick,
    black,
    ]
    table[x=distance,y expr=\thisrowno{2} * 100,col sep=comma]{data/metrics_per_distance_validation.csv};
    \addlegendentry{Validation} 
 
\addplot+[
    mark=none,
    black,
    ]
    table[x=distance,y expr=\thisrowno{2} * 100,col sep=comma]{data/metrics_per_distance_testing.csv};
    \addlegendentry{Testing}   

\end{axis}
\end{tikzpicture}
\caption{Accuracy per edit distance (last epoch)}
\end{subfigure}
\caption{Metrics.}
\label{fig:metrics}
\end{figure}